%% file: main.tex
\pdfoutput=1

\documentclass[11pt]{article}

\usepackage{emnlp2021}
\usepackage{times}
\usepackage{latexsym}
\usepackage{graphicx}
\usepackage{url}

\usepackage[T1]{fontenc}

\usepackage[utf8x]{inputenc}

\usepackage[scaled=.75]{beramono}

\usepackage[
  separate-uncertainty = true,
  multi-part-units = repeat
]{siunitx}

\usepackage{booktabs}
\usepackage{microtype}

\usepackage{mathtools}

\title{VQA-MHUG: A Gaze Dataset to Study Multimodal Neural Attention in Visual Question Answering}

\usepackage{enumitem}
\setlist{topsep=2pt, noitemsep, leftmargin=*}
\usepackage{subcaption}

\usepackage[draft]{changes}

\definechangesauthor[name=Ekta, color=blue]{ekta}

\definechangesauthor[name=Fabian, color=green]{fabian}

\definechangesauthor[name=Florian, color=purple]{florian}

\definechangesauthor[name=Andreas, color=orange]{andreas}

\definecolor{pinkAdnen}{HTML}{C70039}

\definechangesauthor[name=Prajit, color=pinkAdnen]{prajit}

\author{Ekta Sood$^1$, Fabian Kögel$^1$, Florian Strohm$^1$, Prajit Dhar$^2$, Andreas Bulling$^1$\\
$^1$University of Stuttgart, Institute for Visualization and Interactive Systems (VIS), Germany\\ $^2$University of Groningen, Center for Language and Cognition (CLCG), the Netherlands\\
\texttt{\{ekta.sood,fabian.koegel,florian.strohm,andreas.bulling\}@vis.uni-stuttgart.de}\\ 
\texttt{p.dhar@rug.nl}
}

\begin{document}
\maketitle
\begin{abstract}
\input{0_abstract_New}
\end{abstract}

\section{Introduction}
\input{1_introduction}

\section{Related Work}
\input{2_relatedwork}

\input{3_dataset}

\section{Comparison of Human and Machine Attention}
\input{4_methods}

 \input{5_experiments}

\section{Results}
\input{6_analysis_results}

\section{Conclusion and Future Work}
\input{7_conclusion}

\section*{Ethical Statement}
\input{8_ethical_considerations}

\section*{Acknowledgements}
E. Sood was funded by the Deutsche Forschungsgemeinschaft (DFG, German Research Foundation) under Germany's Excellence Strategy - EXC 2075 -- 390740016;
F. Strohm and A. Bulling were funded by the European Research Council (ERC; grant agreement 801708);\\
We would like to especially thank Simon Tannert for his valuable insights and support, as well as Dr. Philipp Müller and Dr. Paul Bürkner for their helpful suggestions. Lastly, we would like to thank the anonymous reviewers for their useful feedback.

\bibliography{anthology,custom}
\bibliographystyle{acl_natbib}

\clearpage
\appendix
\section*{Appendix}
\input{9_appendix}

\end{document}

%% file: 0_abstract_New.tex
We present VQA-MHUG -- a novel 49-participant dataset of \underline{m}ultimodal \underline{hu}man \underline{g}aze on both images and questions during visual question answering (VQA) collected using a high-speed eye tracker.
We use our dataset to analyze the similarity between human and neural attentive strategies learned by five state-of-the-art VQA models: Modular Co-Attention Network (MCAN) with either grid or region features, Pythia, Bilinear Attention Network (BAN), and the Multimodal Factorized Bilinear Pooling Network (MFB). 
While prior work has focused on studying the image modality, our analyses show -- for the first time -- that for all models, higher correlation with human attention on text is a significant predictor of 
VQA performance.
This finding points at a potential for improving VQA performance and, at the same time, calls for further research on neural text attention mechanisms and their integration into architectures for vision and language tasks, including but potentially also beyond VQA.

%% file: 1_introduction.tex
Visual question answering (VQA) has gained popularity as a practically-useful and challenging task at the intersection of natural language processing (NLP) and computer vision (CV)~\cite{Antol.2015}. The key challenge in VQA is to develop computational models that are able to reason over questions and images in order to generate answers that are well-grounded in both modalities~\cite{P.Zhang.2015,agrawal-etal-2016-analyzing,Goyal.2017,Kafle.2019}.
Attention mechanisms originally introduced in NLP for monomodal language tasks have been successfully applied to multimodal tasks (like VQA) and established a new state of the art~\cite{Correia.2021_survey,kim2018bilinear, Yu.2019_mcan}.

These advances have, in turn, triggered research into understanding the reasons for these improvements. A body of work has studied similarities between neural and human attention~\cite{qiuxia2020understanding,yun2013studying,Das.2016}.
Models seem to learn very different attention strategies and similarity to human attention might only improve performance for specific model types~\cite{Sood_2020_Interp}.
However, although VQA is an inherently multimodal task, all of these analyses have only focused on image attention.
The most likely reason for this is that existing datasets only offer mono-modal attention on the image~\cite{Das.2016,fosco2020much,Chen.2020}.
In addition, due to the challenges involved in recording human gaze data at scale, prior works have instead used mouse data as a proxy to attention \cite{Jiang.2015}.
However, mouse data was shown to over-estimate some image areas~\cite{Tavakoli.2017,Das.2016} or to miss relevant background information altogether~\cite{YusukeSugano.2016, Tavakoli.2017b}.
As of now, there is no publicly available dataset that offers human gaze data on both the images and questions.
This severely impedes further progress in this emerging area of research.

Our work fills this gap by introducing VQA-MHUG -- the first dataset of \underline{m}ultimodal \underline{hu}man \underline{g}aze on both images and questions in VQA.
To collect our dataset, we conducted a 49-participant eye tracking study.
We used a commercial, high-speed eye tracker to record gaze data on images and corresponding questions of the VQAv2 validation set.
VQA-MHUG contains 11,970 gaze samples for 3,990 question-image pairs, tagged and balanced by reasoning type and difficulty.
We ensured a large overlap in question-image pairs with nine other VQA datasets to maximize the usefulness of VQA-MHUG for future multimodal studies on human and neural attention mechanisms.
Using our dataset, we conduct detailed analyses of the similarity between human and neural attentive strategies, the latter of which we obtained from five top-performing models in the VQA challenges 2017-2020:
Modulated Co-Attention Network (MCAN) with grid or region features, Pythia, Bilinear Attention Network (BAN), and the Multimodal Factorized Bilinear Pooling Network (MFB). 
These analyses show, for the first time, that correlation with human attention on text is a significant predictor of accuracy for all the studied state-of-the-art VQA models.
This suggests a potential for significant performance improvements in VQA by guiding models to "read the questions" more similarly to humans. 
In summary, our work contributes:
\begin{enumerate}
    \item VQA-MHUG, a novel 49-participant dataset of multimodal human gaze on both \textit{images} and \textit{questions} during visual question answering collected using a high-speed eye tracker.
    \item Detailed analysis of the similarity between human and neural attentive strategies indicating that human-like attention to text could yield significant performance improvements.
\end{enumerate}

%% file: 2_relatedwork.tex
Our work is related to previous work on 1) neural machine attention, 2) attention in VQA, and 3) comparison of neural and human attention.

\paragraph{Neural Machine Attention.}
Inspired by the human visual system, neural machine attention
allows neural networks to selectively focus on particular parts of the input, resulting in significant improvements in performance and interpretability~\cite{Correia.2021_survey}.
Single-modal attention \cite{bahdanau2014neural} as well as approaches that build on it, such as self attention~\cite{xu2015show,vaswani.2017} or stacked attention~\cite{yang2016stacked,yang2016hierarchical,Zhang.2018_learn_count,anderson2018bottom}, have been shown to be particularly helpful for sequence learning tasks in NLP and CV.
Initially, attention mechanisms were often combined with recurrent and convolutional architectures to encode the input features~\cite{bahdanau2014neural,yu2017multi,Tavakoli.2017,Kim.05.06.2016,Lu.2016,Jabri.2016,agrawal-etal-2016-analyzing}.
More recently, Transformer-based architectures have been introduced that solely rely on attention~\cite{vaswani.2017,Yu.2019_mcan,Khan.27.10.2020}.
Large-scale, pre-trained language models
are a key application of Transformers that enabled their current performance lead in both NLP and multimodal vision-language tasks~\cite{devlin2018bert,yang2019xlnet,Yu.2019_mcan,lu2019vilbert}.

\paragraph{Attention in VQA.}
Increased interest into capturing multimodal relationships with attention mechanisms have put focus on VQA as a benchmark task~\cite{Malinowski_Multi_2014,malinowski2015ask,Lu.2016,yu2017multi,nguyen2018improved,yang2019co,li2019beyond}.
In fact, attention mechanisms have been extensively explored in VQA and have repeatedly dominated the important VQAv2 challenge~\citep{anderson2018bottom,Yu.2019_mcan,Jiang.2020_grid_feat}.
Although attention-based models have achieved remarkable success, it often remains unclear how and why different attention mechanisms actually work~\cite{jain2019attention,serrano2019attention}.
\paragraph{Comparing Neural and Human Attention.}
Several prior works have proposed datasets of human attention on images to study the differences between neural and human attention in VQA~\cite{Das.2016,fosco2020much,Chen.2020}.
In particular, free-viewing and task-specific mouse tracking from SALICON~\cite{Jiang.2015} and VQA-HAT~\cite{Das.2016}, as well as free-viewing and task-specific gaze data from SBU Gaze~\cite{Yun.2015} and AiR-D~\cite{Chen.2020} have been compared to neural attention. 
All of these works were limited to images only and found mouse tracking to overestimate relevant areas and miss scene context~\cite{YusukeSugano.2016, Tavakoli.2017, Tavakoli.2017b, He.2019}.
Furthermore, while integrating human attention over the image showed performance improvements in VQA~\cite{Park_2018_CVPR, Qiao.2018,Chen.2020}, the influence of integrating human text attention remains unclear.

There is currently no multimodal dataset including real human gaze on VQA questions and images.
This represents a major limitation for two different aspects of research, i.e. research aiming to better understand and improve neural attention mechanisms and research focusing on integrating human attention to improve VQA performance.

%% file: 3_dataset.tex
\section{The VQA-MHUG Dataset}
We present \underline{V}isual \underline{Q}uestion \underline{A}nswering with \underline{M}ulti-Modal \underline{Hu}man \underline{G}aze (VQA-MHUG)\footnote{The dataset is publicly available at \url{https://perceptualui.org/publications/sood21_conll/}}. 
To the best of our knowledge, this is the first resource containing multimodal human gaze data over a textual question and the corresponding image. 
Our corpus encompasses task-specific gaze on a subset of the benchmark dataset VQAv2 val\footnote{\url{https://visualqa.org/download.html}}~\cite{goyal2017making}. 
We specifically focused on question-image pairs that machines struggle with, but humans answer easily (determined by high inter-agreement and confidence in the VQAv2 annotations).
We then balanced the selection by evenly picking questions based on a machine difficulty score and from different reasoning types.
Thus, VQA-MHUG covers a wide range of challenging reasoning capabilities and overlaps with many VQAv2-related datasets (see Table \ref{tab:mhug-dataset-overlap} in Appendix \ref{sec:VQA-MHUG_Overlap}).

\paragraph{Reasoning Types.}
VQAv2 groups question-image pairs based on question words: \textit{what}, \textit{who}, \textit{how}, \textit{when} and \textit{where}. Instead, we binned our pairs into the reasoning capabilities required to answer them. We incorporated the categories proposed by~\citet{Kafle.2017} for their task directed image understanding challenge (TDIUC) and extended them with an additional category, \textit{reading}, for questions that are answered by reading text on the images. This resulted in 12 reasoning types that align better with commonly-diagnosed error cases\footnote{See Appendix \ref{sec:bins} for details on the reasoning type tagging.}. We binned VQAv2 val pairs accordingly by training a LSTM-based classifier on 1.6\,M TDIUC and 145\,K VQAv2 train+val samples which we labelled using regular expressions. The classifier predicted the reasoning type for a given question-answer pair. The final model achieved 99.67\% accuracy on a 20\% held-out test set.

\paragraph{Machine Difficulty Score.}
To assess the difficulty for a machine to answer a question-image pair, we ran two popular VQA models -- MFB~\cite{yu2017multi} for multimodal fusion and MCAN~\cite{Yu.2019_mcan} for transformer attention -- inspired by~\citet{Sood_2020_Interp}.
A difficult question results in low answer accuracy, particularly after rephrasing or asking further control questions.
To test this, we evaluated on four datasets and averaged their corresponding normalized metrics: (1) VQAv2 accuracy, (2) VQA-CP accuracy on reduced bias~\cite{agrawal2018don}, (3) VQA-Introspect's consistency with respect to visual perception~\cite{selvaraju2020squinting} and (4) VQA-Rephrasings' robustness against linguistic variations~\cite{shah2019cycle} (see Appendix \ref{sec:diffscores}).

\paragraph{Participants and Experimental Setup.}
We recruited 49 participants at the local university (18 identified female and 31 male) with normal or corrected-to-normal vision, aged between 19 and 35 years ($\mu=25.8$, $\sigma=2.8$) and compensated them for their participation\footnote{The university ethics committee approved our study.}. 
All participants had an English Level of C1 or above (8 were native speakers).\footnote{After providing their consent, we collected basic demographic information for each participant. The anonymized data is available with the dataset.}

Questions and images were presented one after each other on a 24.5" monitor with resolution 1920x1080\,px. 
They were centered on a white background, and scaled/line-wrapped to fit 26.2x11.5° of visual angle in the center. 
For the questions, we used a monospace font of size 0.6° and line spacing such that the bounding boxes around each word covered 1.8° vertically. 
Binocular gaze data was collected with an EyeLink 1000 Plus remote eye tracker at 2 kHz with an average measured tracking error of 0.62° (see Appendix \ref{sec:experimental_setup}).

Participants had unlimited viewing time but were instructed to move on as soon as they understood the question, gave an answer, or decided to skip.
They completed a set of practice recordings to familiarize themselves with the study procedure.
As such, the task was known to the participant, so both the question reading and the subsequent image viewing were conditioned on VQA.
They then completed three blocks of 110 recordings in randomized order with 5 minute breaks in-between.

\paragraph{Dataset Statistics.}
VQA-MHUG contains gaze on 3,990 stimuli from VQAv2 val.
For each stimulus, we provide three recordings from different participants over text and image, their corresponding answer, and whether they answered the question correctly (as compared to the VQAv2 annotations).
For 3,177 stimuli (79.6\%), the majority of participant answers appear in the VQAv2 annotations.

\paragraph{Human Attention Maps.}
To generate human attention maps, we used the fixation detection algorithm of the EyeLink software with default parameters.
We always picked the eye with the lower validation error to prioritize accuracy \cite{Hooge.2019} and represented fixations by Gaussian kernels with $\sigma = 1\text{°}$.
For our experiments, we assumed that the majority of gaze is valid and averaged the three recordings per stimulus, yielding a single attention map.  

\begin{figure}[!t]
    \begin{subfigure}{0.49\linewidth}
        \centering
        \includegraphics[width=\textwidth]{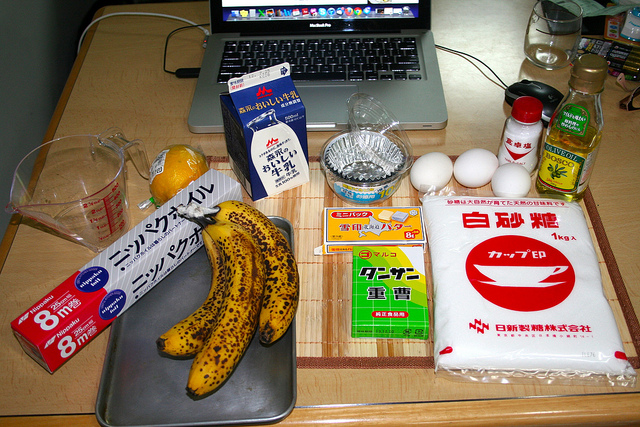}
        \caption{collection study}
        \label{fig:validation_original}
    \end{subfigure}
    \begin{subfigure}{0.49\linewidth}
        \centering
        \includegraphics[width=\textwidth]{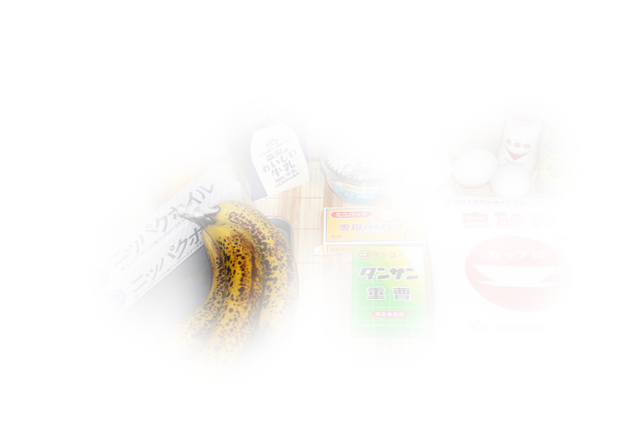} 
        \caption{validation study}
        \label{fig:validation_masked}
    \end{subfigure}
    \caption{Example images for the question "How ripe are the bananas?". Validation images (b) were masked using the attention maps from our VQA-MHUG dataset.}
    \label{fig:validation}
\end{figure}

\paragraph{Dataset Validation.}
To validate that the attention maps indeed contain relevant image regions, we masked 300 stimuli with our recorded VQA-MHUG maps (see Figure \ref{fig:validation_masked}). 
Then, we showed two additional participants these masked stimuli.
Comparing their answer accuracy with the participants who saw the full images, validation participants achieved comparable accuracy (62.43\% vs. 63.87\% in the main study). 
Therefore our VQA-MHUG maps contain sufficient image areas to answer the questions and mask distracting objects as illustrated in Figure \ref{fig:validation}.

\paragraph{Comparison to Related Datasets.}
\label{sec:related_datasets}
We further measured the center bias and compared VQA-MHUG to related human attention datasets~\cite{Jiang.2015,Das.2016,Chen.2020} on their overlapping samples. 
All datasets use mouse tracking as a proxy to collect human attention, except for the eye-tracking dataset AiR-D~\cite{Chen.2020} which is similar to our recording paradigm, yet has no overlap with VQAv2.
Therefore, we showed participants 195 additional stimuli from the AiR-D dataset for comparison. 
Table \ref{tab:mhug-dataset-corr} shows the mean rank correlation
of VQA-MHUG with a synthetic center fixation, inter-participant, and the other datasets.
The high correlation between VQA-MHUG and AiR-D indicates that our data is of comparable quality. 
Our center bias is smaller compared to AiR-D but, as expected from human eye behaviour \cite{Tatler2007TheCF}, larger than in the mouse tracking proxies SALICON and VQA-HAT.
We observe that both mouse tracking datasets have significantly lower correlation with VQA-MHUG than the eye-tracking AiR-D corpus.

\begin{table*}
\centering
  \begin{tabular}{l r l l} 
    \toprule
    && \multicolumn{1}{c}{\textbf{VQA-MHUG}} & \multicolumn{1}{c}{\textbf{Center Fixation}}\\
    \midrule
    \textbf{Dataset} & \textbf{Method} &
    \textbf{{\boldmath$\rho$} $\uparrow$} 
    &\textbf{{\boldmath$\rho$} $\uparrow$} 
    \\
    \midrule
    
    VQA-MHUG & G 
    & $0.769\pm 0.079$ 
    & $0.473\pm 0.049$ 

    \\
    AiR-D & G 
    & $0.710\pm 0.060$ 
    & $0.523$ 

    \\
    VQA-HAT & M 
    & $0.612\pm 0.145$
    & $0.339\pm 0.107$ 
    
    \\

    SALICON & M
    & $0.634\pm 0.063$
    & $0.479$ 
    \\
    \bottomrule
    \multicolumn{4}{l}{\footnotesize{\textbf{G}: Gaze, 
    \textbf{M}: Mouse-Tracking
    }}
  \end{tabular}
 
  \caption{Spearman's rank correlation ($\rho$) 
  of VQA-MHUG with itself (inter-participant), related datasets, and a synthetic center fixation -- Mean over all samples in the intersection of the datasets and three VQA-MHUG participants. The standard deviation is the mean error over participants.
  Only VQA-HAT and VQA-MHUG provide multiple attention maps per sample, allowing us to calculate the standard deviation when comparing to the synthetic center fixation.
  }
  \label{tab:mhug-dataset-corr}
\end{table*}

%% file: 4_methods.tex
The collected data enabled us to analyze whether models achieve a higher accuracy on VQAv2 val the more their attentive behavior over the text and image correlates with human ground-truth attention.
Hence, we investigated the attention weights over text and image features of different SOTA VQA models.

\subsection{VQA Models}
We selected five top performing VQA models of the VQA challenges 2017 to 2020:
\begin{itemize}
    \item MFB~\cite{yu2017multi} (Runner-up 2017);
    \item BAN~\cite{kim2018bilinear} (Runner-up 2018);
    \item Pythia v0.1~\cite{jiang2018pythia} (Winner 2018);
    \item MCAN\textsubscript{R} with region image features ~\cite{Yu.2019_mcan} (Winner 2019);
    \item MCAN\textsubscript{G} with grid image features~\cite{Jiang.2020_grid_feat} (Winner 2020).
\end{itemize}
Instead of using the text and image features directly for classification, these models re-weight the features using linear, bilinear and Transformer~\cite{vaswani.2017} (co-)attention mechanisms, whose attention maps we extracted and compared to human ground-truths from VQA-MHUG.

Pythia and MFB use co-attention: they first use a projected attention map to re-weight text features, then fuse them with the image features using linear (Pythia) and bilinear (MFB) fusion and subsequently re-weight the image features using an attention map projected from the fused features.
In this way, the text attention influences the image attention. 
BAN avoids separating the attention into text and image streams and reduces both input streams simultaneously with a bilinear attention map projected from the fused features.
Finally, MCAN as a Transformer model stacks co-attention modules with multi-headed scaled dot-product attention for each modality.
After the last Transformer layer in both the text and image stream, another attention map is used to project the feature matrix into a single feature vector.

\subsection{Extracting Model Attention}
We used
an official implementation\footnote{\url{https://github.com/zwxalgorithm/pythia}} of the Pythia v0.1 architecture and the OpenVQA\footnote{\url{https://github.com/MILVLG/openvqa}} implementations~\cite{yu2019openvqa} for MFB, BAN and MCAN.
We re-implemented the grid image feature loader for MCAN\textsubscript{G}, since it is not available in OpenVQA.

Following previous work~\cite{Sood_2020_Interp}, we trained each network architecture twelve times with random seeds on the VQAv2 training set and then chose the top nine models based on the validation accuracy. 

For models based on region image features, we used the extracted features provided by Anderson et al.~\citeyear{anderson2018bottom}, while we trained MCAN\textsubscript{G} with ResNeXt~\cite{xie2017aggregated} grid features as provided by the authors~\cite{Jiang.2020_grid_feat}\footnote{\url{https://github.com/facebookresearch/grid-feats-vqa}}.

For MFB and Pythia we extracted the two projected attention maps that re-weight text and image features, while we extracted the single bilinear attention map for BAN.
To obtain separate attention maps for text and image from BAN's bilinear attention map, we marginalized over each dimension as suggested by the authors~\cite{kim2018bilinear}.
MFB, BAN and Pythia generate multiple such attention maps called ``glimpses''
by using multiple projections.
We averaged the glimpses after extraction, yielding a single attention map for each modality.
Since it is unclear how the Transformer layer weights relate to the original input features, we instead extracted the attention weights of the final projection layer in text and image streams for MCAN\textsubscript{R} and MCAN\textsubscript{G}.

The extracted image attention maps contain one weight per feature. 
To compare them with the human spatial attention maps collected in VQA-MHUG, we mapped the features back to their source region in the image. 
For region-based features we assigned the attention weights to the corresponding bounding box normalized by region size.
Analogously, for grid-based features, we mapped the attention weights to their corresponding grid cells.
The text attention vector was directly mapped back to the question token sequence.
We excluded 74 samples due to varied tokenization between models.

%% file: 5_experiments.tex
\begin{figure*}[ht]
    \begin{subfigure}{0.374\linewidth} 
        \centering
        \includegraphics[width=\textwidth]{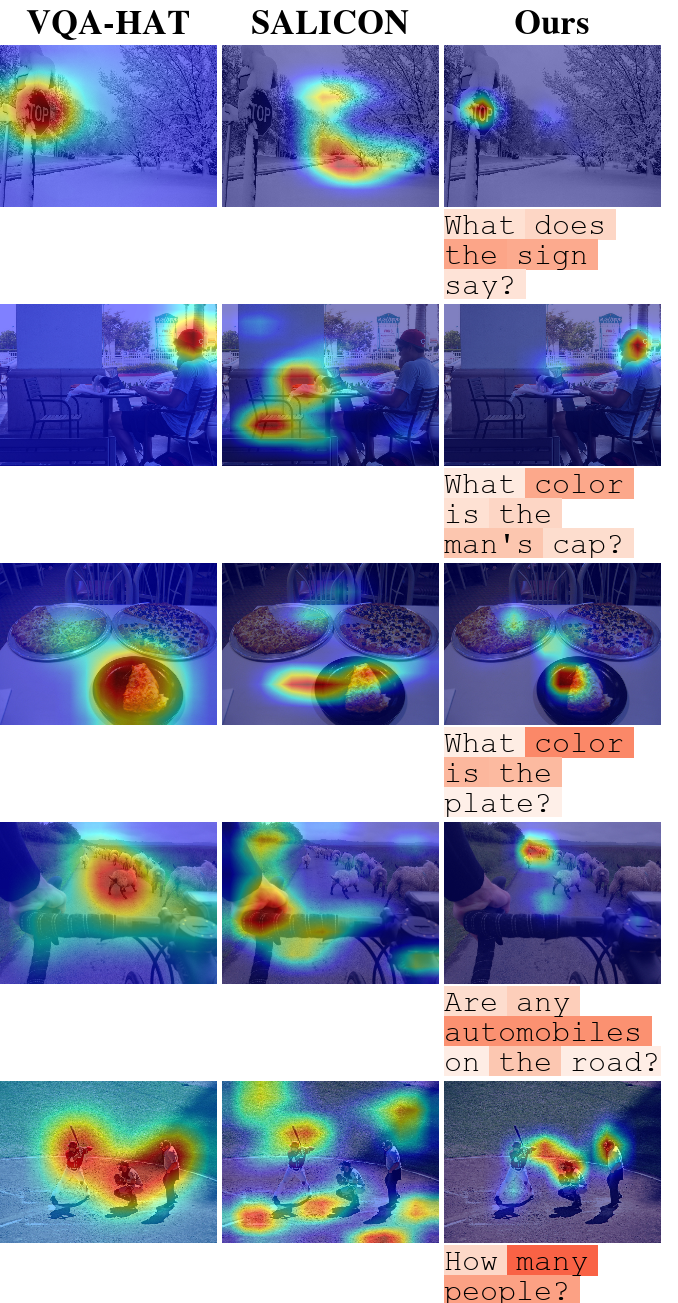}
        \caption{comparison to other attention datasets}
        \label{fig:visualization_datasets}
    \end{subfigure}
    \begin{subfigure}{0.624\linewidth} 
        \centering
        \includegraphics[width=\textwidth]{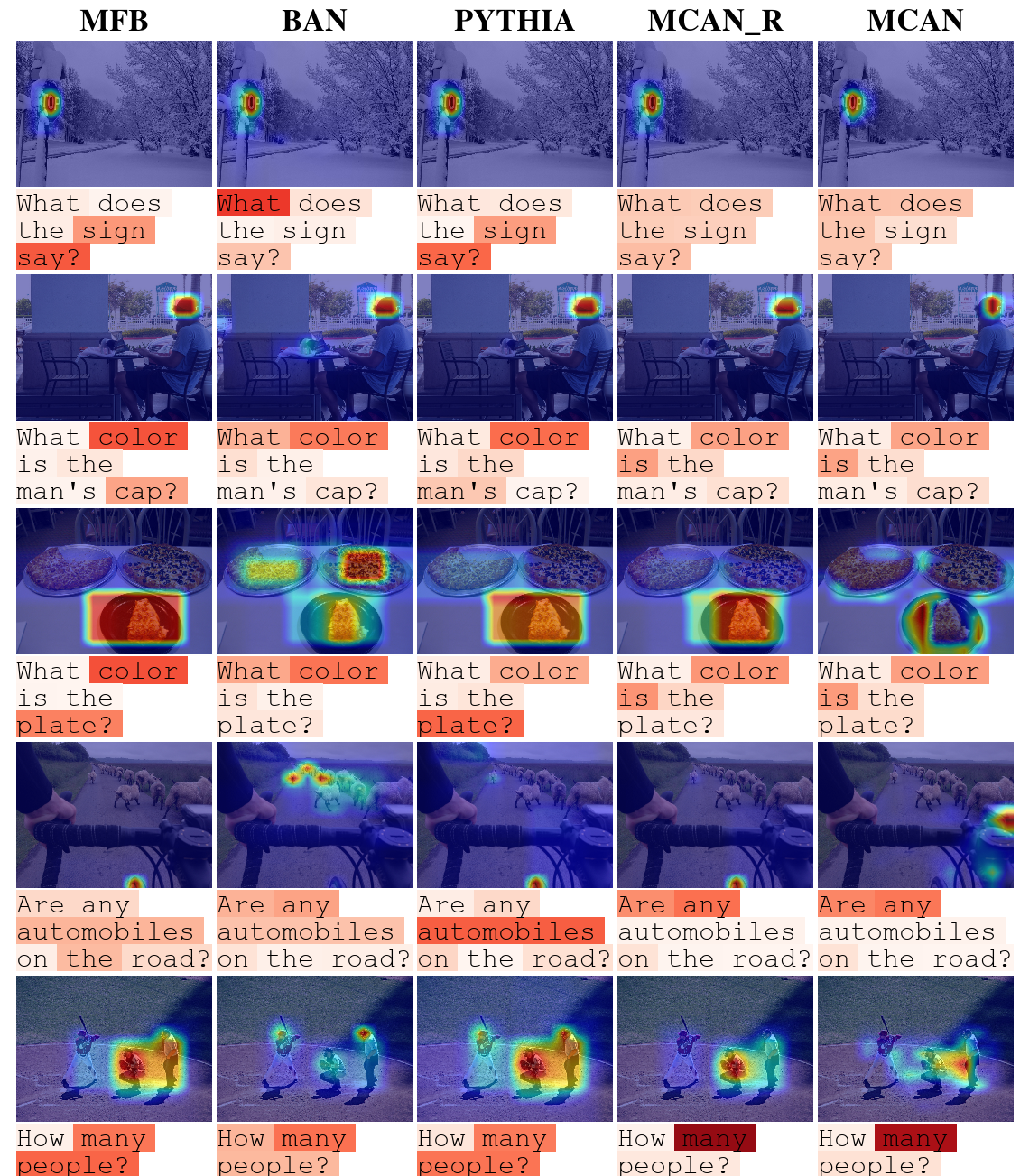} 
        \caption{model attention - text, image, and inter-modal comparison}
        \label{fig:visualization_models}
    \end{subfigure}
    \caption{Attention maps visualized across question types.
    Image attention seems mostly plausible throughout models. Previous datasets lack attention on the questions, but we reveal now that text attention is not always human-like, nor plausible. Mouse tracking datasets, SALICON and VQA-HAT, seem to over-estimate the relevant areas.
    }
    \label{fig:visualization}
\end{figure*}

\subsection{Performance Metrics}
\label{experiments:metrics}
We compared the multimodal attention extracted from five models to our human data in VQA-MHUG using three approaches.
We used Spearman's rank correlation to compare importance ranking of image regions and words, Jensen-Shannon divergence to compare the distance between the human, and neural attention distributions 
and a regression model to study the suitability of text and image correlation as predictors of per document model accuracy.

\paragraph{Spearman's rank correlation and Jensen-Shannon divergence.} Similar to prior work, we downsampled all attention maps to 14x14 matrices and calculated the mean \textit{Spearman's rank correlation} $\rho$~\cite{Das.2016} and \textit{Jensen-Shannon divergence} (JSD)~\cite{Sood_2020_Interp,Sood_2020_Improve} between the neural attention and the corresponding human attention. 
We computed both metrics for both image and text modalities.
We also evaluated the corresponding accuracy scores on the VQAv2 validation set~\cite{VQAEval}.

\paragraph{Ordinal Logistic Regression.}
Averaging correlation over the whole dataset is too coarse and obscures the impact that similarity to human attention has on accuracy.
Additionally, rank correlation does not allow to analyze the effect of two independent variables on a dependent variable~\cite{Bewick2003StatisticsR7}, e.g. image and text attention correlation on accuracy.
To account for this and to study on a per document basis which modality factors influence the likelihood of a model to predict the answer correctly, we performed an \textit{Ordinal Linear Regression} (OLR).

The official VQAv2 evaluation~\cite{VQAEval} score per document is based on agreement with ten human annotator answers, where each match increases the score by 0.3 (capped at 1.0 or 4 agreed answers).
Since our response variable (accuracy score) is not necessarily ordered equidistant, we binned accuracy scores for each document into a likelihood scale (accuracy correctness).

The model predicts the likelihood of accuracy correctness for each document with three different predictors — the text correlation ($x$), the image correlation ($y$), and the interaction between the text and image correlation ($z$). The latter we deem inter-modal correlation predictor, as it allows us to test if the interaction between the correlation of text and image impacts accuracy. Given that the dependant variables are ranked we opted for using ordered logistic regression to predict for each accuracy bin.

%% file: 6_analysis_results.tex
\subsection{Human and Neural Attention Relationship -- Averaged Over Documents}
Table~\ref{tab:inter-model-avg} shows the overall accuracy scores of the five models on the VQAv2 validation set when trained only on the training partition. The models improved over the challenge years -- MCAN grid is the current SOTA~\cite{Jiang.2020_grid_feat}. 
For each model and modality, we report the Spearman's rank correlation and JSD scores averaged over the entire VQA-MHUG corpus (cf. Section~\ref{experiments:metrics}).
All figures were averaged over nine model runs and the standard deviation is given over those instances.
Given that one cannot average p-values we used a paired t-test to check if the differences in correlation and JSD per document and between models were statistically significant at $p<0.05$ (see Appendix \ref{sec:test_signf_between_models}). 

\paragraph{Image attention.} 
Models using region features, i.e. excluding MCAN$_{\text{G}}$, are more correlated with human visual attention on images.
MCAN$_{\text{R}}$ achieves the highest correlation, MFB the lowest, and the general trend shows that models with higher correlation had higher overall validation accuracy.
Although MCAN$_{\text{G}}$ achieves the highest accuracy, it had the lowest correlation with human image attention. 
For all model types, the difference between image correlation scores is significant, except between Pythia and BAN (see Appendix \ref{sec:test_signf_between_models}). 
With respect to the JSD, we observed similar patterns except for the Pythia model, which was more dissimilar to human attention (had a higher overall JSD) compared to BAN. 
For all model types, the difference between image JSD scores was statistically significant (see Appendix \ref{sec:test_signf_between_models}). 

\paragraph{Text attention.} 
Both the correlation and JSD scores indicate that Pythia is the most similar to human text attention, followed by MFB.
Models with higher overall accuracy do not have high similarity to human visual attention over text on the JSD and correlation metrics. 
For both metrics, the difference in text attention between every model pairing is statistically significant, except for the JSD scores between pairings of BAN, MCAN$_{\text{G}}$, and MCAN$_{\text{R}}$ (see Appendix \ref{sec:test_signf_between_models}).

\begin{table*}
\centering
  \resizebox{\textwidth}{!}{\begin{tabular}{l l l l l l } 
    \toprule
    & & \multicolumn{2}{c}{\textbf{Image}} & \multicolumn{2}{c}{\textbf{Text}}\\
    \midrule
    \textbf{Model} & \textbf{Accuracy} & \textbf{{\boldmath$\rho$} $\uparrow$} 
    & \textbf{JSD $\downarrow$}&\textbf{{\boldmath$\rho$} $\uparrow$}
    & \textbf{JSD $\downarrow$}\\
    \midrule
    MCAN$_{\text{G}}$   & 70.24\%   & \SI{0.509 \pm 0.026}  
    & \SI{0.537 \pm 0.003} &                  \SI{-0.059 \pm 0.012} 
    & \SI{0.402 \pm 0.007}\ \\
    MCAN$_{\text{R}}$   & 67.24\%   & \SI{0.602 \pm 0.003}  
    & \SI{0.467 \pm 0.002} &                  \SI{-0.042 \pm 0.018}  
    & \SI{0.398 \pm 0.017}\ \\
    Pythia              & 66.00\%   & \SI{0.584 \pm 0.003}     
    & \SI{0.479 \pm 0.001} & \hspace{0.53em}  \SI{0.251 \pm 0.016}   
    & \SI{0.337 \pm 0.015}\ \\
    BAN                 & 65.91\%   & \SI{0.582 \pm 0.004} 
    & \SI{0.469 \pm 0.002} &                  \SI{-0.132 \pm 0.030}    
    & \SI{0.398 \pm 0.021}\ \\
    MFB                 & 65.06\%   & \SI{0.530 \pm 0.003}  
    & \SI{0.523 \pm 0.004} & \hspace{0.53em}  \SI{0.225 \pm 0.055}    
    & \SI{0.352 \pm 0.011}\ \\
    \bottomrule
  \end{tabular}}
  \caption{Accuracy of the five models as well as the Spearman's rank correlation ($\rho$) and the Jensen–Shannon divergence (\textit{JSD}) between neural and human attention over images (left) and text (right). Standard deviation was calculated over nine model runs and indicates the attention variability between different instances of the same architecture. All correlation and JSD scores between models differ significantly (p<0.05), except for the image correlation between Pythia and BAN as well as the JSD text scores between BAN, MCAN$_{\text{G}}$ and MCAN$_{\text{R}}$}
  \label{tab:inter-model-avg}
\end{table*}

\subsection{Ordinal Logistic Regression} 
By averaging evaluation metrics (correlation and JSD) across documents, we obscure the impact that similarity has \textit{on each document} with respect to accuracy. The Ordinal Logistic Regression model results uncover the importance of the text and image correlation scores as predictors on per document accuracy. 

\paragraph{Text Correlation.}
We show (cf. Table~\ref{tab:linear_regression}) \textbf{for all five different VQA models}, that as the correlation to human text attention decreases, the likelihood that the models will be able to correctly predict the answer significantly decreases/ Our findings show that correlation to human text attention is a significant predictor on accuracy.
The MCAN$_{\text{G}}$, MCAN$_{\text{R}}$, and MFB model have the strongest relationship ($p$ < 0.001) to text correlation being a significant predictor on accuracy. This indicates that for these models in particular, the less the model is correlated with human text attention, the less likely the model will predict the answer correctly. 

\paragraph{Image Correlation.}
Interestingly, we observe the same trend as text correlation, in which image attention correlation is also a significant predictor on accuracy, but \textbf{not consistently across all models}.
It is a significant predictor for three (MCAN$_{\text{G}}$, Pythia, and BAN) out of the five total models. Notably, the MCAN$_{\text{G}}$ model has a significantly strong relationship to image correlation. 
This indicates that when the Pythia, BAN, and in particular MCAN$_{\text{G}}$ learn attention which is less correlated to human image attention, then the model is more less likely to be able to predict the answer correctly.

\paragraph{Inter-Modal Correlation.} We paired the text correlation $x$ and the image correlation $y$ together as an inter-modal predictor $z$. Inter-modal correlation tests whether the interaction between the two correlation scores, as the predictor $z$, has an effect on accuracy. 
Interestingly, inter-modal correlation $z$ is a significant predictor on accuracy for the MCAN$_{\text{G}}$ and Pythia models but not for the other 3 model types.

\paragraph{Attention Maps --  Qualitative Analysis.}
Figure \ref{fig:visualization} visualizes the human as well as neural attention distributions of five VQA models for a selection of examples from the benchmark VQAv2 dataset.\footnote{See Appendix \ref{sec:more_examples} for additional examples.}
As can be seen, all previous datasets only uncover the differences between human and neural image attention, while VQA-MHUG (ours) allows for studying multimodal neural VQA models attention. 
We also find our attention maps to be highly relevant and confirm that the mouse tracking datasets SALICON and VQA-HAT seem to over-estimate relevant areas. 
As the AiR-D dataset does not overlap with VQAv2, we separately visualize a selection of examples (see Section~\ref{sec:related_datasets} overlapping with our VQA-MHUG data (see Appendix \ref{sec:mhug-vs-air}). 

\subsection{Discussion}
When averaging metrics across all documents in VQA-MHUG, our results regarding similarity between machine and human image attention and performance follow insights derived from previous work~\cite{Das.2016}, where they observed that as the models improved with respect to accuracy they were also more correlated to human attention on the images. However, notably we only observe this trend with the models which use region features. That is, though the MCAN grid is the highest performing model with respect to accuracy, it is also the model which is least similar to human image attention. Such an observation was also reported in previous work which compared the XLNet transformer to human attention~\cite{Sood_2020_Interp}.

Analysis from the Ordinal Logistic Regression model shows, for the first time, that correlation to human text attention is a significant predictor across all VQA model types, where dissimilarity between human and neural text attention decreases the likelihood of the models ability to predict the answer correctly. 
We conclude that striving to enhance neural attention to more similarly emulate human attention on text will improve performance in the five VQA models.
As can be observed in Figure \ref{fig:visualization}, text attention is not always human-like, especially for the otherwise high performing MCAN models, suggesting that increased similarity to human text attention might lead to further improvements with respect to accuracy.

Due to the lack of human attention data over text, researchers were not able to uncover the limitations or relevance of high correlation to human text attention on VQA model accuracy.
In addition, our analysis on the role of image attention and inter-modal attention as a predictor on accuracy indicates that for certain model types it would be beneficial to improve image and inter-modal correlation. 
These findings are consistent with~\citet{Sood_2020_Improve} which found that different model types learn different attention strategies and similarity of machine to human attention does not guarantee best performance. This may be due to factors such as features used (grid versus region), the different learned attention strategies across model types and how the architectures model the interactions between the multimodal input features. For example, the MCAN grid model applies self- and guided attention to model the interplay between grid-based image and text feature representations. On the other hand, the Pythia model uses both bottom up attention (image features extracted on the region level) and top down attention (text attention applied over the images), where the text attention weights are not learned by the image feature representations. 

\begin{table}
\centering
\begin{tabular}{@{}llll@{}}
\toprule
\textbf{Model}    & \textbf{Text}   & \textbf{Image}  & \textbf{Inter-Modal}     \\ 
\midrule
MCAN$_{\text{G}}$ & -4.60***        & -8.32***        & -8.33***        \\
MCAN$_{\text{R}}$ & -5.50***        & \phantom{-}0.21 & \phantom{-}0.05 \\
Pythia            & -2.83**         & -1.83*          & -1.81*          \\
BAN               & -2.20*          & -3.62***        & \phantom{-}0.53 \\
MFB               & -3.32***        & -0.05           & -0.208          \\ 
\bottomrule
\end{tabular}

\caption{Ordinal Logistic regression model t-values 
 such that for every one unit decrease in correlation, each respective model is less likely to predict the answer correctly. Significance is denoted as *$p<0.05$, **$p<0.01$, ***$p<0.001$. 
Correlation to human text attention is a significant predictor of accuracy for \textbf{all five models}. Correlation to human image attention is an important for the accuracy of MCAN$_{\text{G}}$, Pythia, and BAN 
 while  inter-modal correlation
 is a significant predictor of accuracy for both MCAN$_{\text{G}}$ and Pythia.}
\label{tab:linear_regression}
\end{table}

%% file: 7_conclusion.tex
In this work we have presented VQA-MHUG -- a new, fully annotated 49-participant dataset for visual question answering that includes nearly 4,000 question-answer pairs.
Our dataset is unique in that it is the first to provide real human gaze data on both images and corresponding questions and, as such, allows researchers to jointly study human and machine attention.
Revealed through a detailed comparison of multiple leading VQA models, we showed that higher correlation between neural and human text attention is a significant predictor of high VQA performance.
This novel finding highlights the potential to improve VQA performance with human-like attention biases and simultaneously calls for further investigation of neural text attention mechanisms, as we find these are an indicator for success on language and vision tasks, including VQA.

%% file: 8_ethical_considerations.tex
We identified a number of potential benefits and risks of our approach.

\paragraph{Potential benefits} By leveraging human behavioral data, our method could be used to guide intelligent user interfaces using human attentive abilities within the context of reading behaviors.
We see significant potential of approach to interpret text attention to enable a new generation of attentive text interfaces, particularly when jointly modelling with user task specific eye movement behaviors during comprehension tasks. We see potential for e-learning multimodal applications approach could be used to qualify reader actions and provide feedback to encourage improvement in comprehension. By bridging the gap between human and neural attention, we see a potential positive impact in improving attention strategies in users. 

\paragraph{Potential risks} Though we see the aforementioned potential benefits, we also identified a some risks and ethical concerns.
By aiming to interpret the gap between human and machine attention, we open the door for potentially exploiting user biases. In addition, one can conceive that there is potential for using the findings of our work to develop tool which discriminate against specific users given their eye movement behaviors. This leads to the discussion
about the behavioral data collection, it is conceivable that one could generate a system which might predicts cognitive impairments in order to filter out individuals from some program or opportunity. 

\paragraph{Dataset Curation}
To protect the privacy of our participants we saved all data anonymized and collected only directly relevant data and demographic information in compliance with our university's code of ethics and the General Data Protection Regulation (GDPR) of the European Union (EU). Our study was approved by the ethics committee (institutional review board) of the university. Additional measures for safety during the COVID-19 pandemic were taken with disinfection of the material, obligatory masks and breaks between scheduled recording sessions. All participants signed a consent form that included details about the purpose, goal, procedure, risks, benefits and privacy measures of our research. For the 45-60 minute study an above average compensation of 20€ was paid. At any point the participant could abort the study without penalty. The study took place in a standard university lab and the participant's head was not fixed. Every 15 minutes a 5 minute break was scheduled.

%% file: 9_appendix.tex
\section{VQA-MHUG Overlap to Related Datasets}
\label{sec:VQA-MHUG_Overlap}
During the selection of stimuli for VQA-MHUG, we maintained large overlaps with other benchmark and attention datasets that also used subsets of VQAv2 questions/images to allow for easy integration and comparison of our data with existing approaches (see Table \ref{tab:mhug-dataset-overlap}). 
\begin{table}[!ht]
\centering
  \begin{tabular}{l l} 
    \toprule
    \textbf{Dataset} & \textbf{∩}\\
    \midrule
        VQAv2 val \cite{Goyal.2017}& 3,990\\
        VG \cite{Krishna.2017}& 2,238\\
        VQA-CP2 \cite{Agrawal.2018}& 1,904\\
        VQA-Rephrasings \cite{Shah.2019}& 1,373\\
        VQA-Introspect \cite{Selvaraju_2020_CVPR} & 1,213\\
        SALICON \cite{Jiang.2015}& 1,134\\
        TDIUC \cite{Kafle.2017}& 1,125\\
        VQS \cite{Gan.2017}& 695\\
        VQA-X \cite{Park_2018_CVPR} & 491\\
        VQA-HAT \cite{Das.2016}& 410\\
    \bottomrule
  \end{tabular}
  \caption{Overlap of VQA-MHUG question-image pairs with different established VQA related datasets.}
  \label{tab:mhug-dataset-overlap}
\end{table}

\section{Reasoning Types}
\label{sec:bins}

\begin{figure*}[ht]
    \centering
    \includegraphics[width=\textwidth]{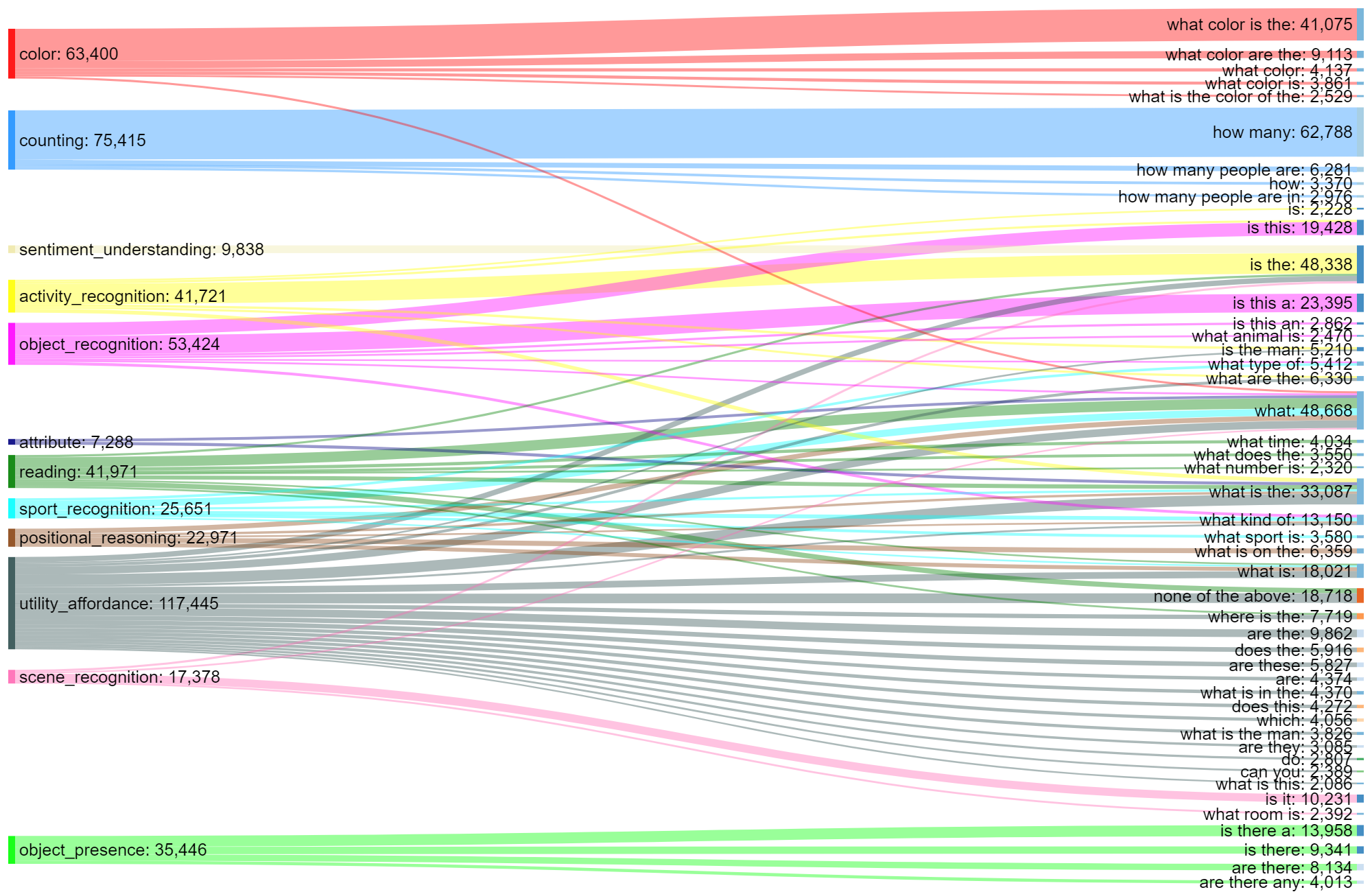}
    \caption{Relationship of our VQA-MHUG reasoning types (left) and VQAv2 question types
(right). Question types are not good categories for error case analysis since they mix many
reasoning capabilities.}
    \label{fig:rtypes_qtypes}
\end{figure*}

We binned question-image pairs by 12 reasoning types, as they align better with potential error classes than the VQAv2 question types. Figure \ref{fig:rtypes_qtypes} shows the relationship of reasoning types to question types. The reasoning types incorporate the categories proposed by \citet{Kafle.2017}, except the absurd category and adding a new \textit{reading} category for questions that ask about text on the images.

\begin{itemize}
    \item Scene Recognition
    \item Object Presence
    \item Colour
    \item Positional Reasoning
    \item Counting
    \item Utility Affordance
    \item Object Recognition
    \item Activity Recognition
    \item Attribute
    \item Reading
    \item Sentiment Understanding
    \item Sport Recognition
\end{itemize}

\subsection{Tagger}
\begin{figure}[ht]
    \centering
    \includegraphics[width=\linewidth]{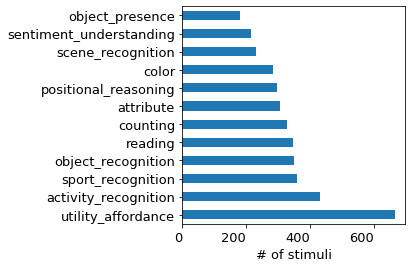}
    \caption{Final distribution of tagged reasoning types in VQA-MHUG. When no other type fit, the tagger assigned \textit{utility affordance}, which had the least training data. This indicates that there could be clusters that do not fit any current type.}
    \label{fig:rtypes_in_mhug}
\end{figure}

To label VQA-MHUG with our reasoning types we used a LSTM-based classifier to predict the reasoning type given the question-answer pair. 
The input text is encoded using 300D glove embeddings~\cite{pennington-etal-2014-glove}, which are passed though a single LSTM layer with hidden size 256 and a final softmax classification layer. 
We labeled 145\,K VQAv2 train-val questions and extended the 1.6\,M TDIUC questions by the \textit{reading} category using regular expressions and manual work. 
We trained the network using this data by optimizing cross-entropy loss with the Adam optimizer and a batch size of 128. 
The final model achieves an accuracy of 99.67\% on a held-out set of 20\% of the training data.
The trained tagger was then used to label the question-image pairs in VQA-MHUG.
Figure \ref{fig:rtypes_in_mhug} shows the label distribution.

\section{Machine Difficulty Score}
\label{sec:diffscores}
For our machine difficulty score we evaluated the Multimodal Factorized Bilinear Pooling Model (MFB) \cite{yu2017multi} for multimodal fusion and the Multimodal Co-Attention Network (MCAN) \cite{Yu.2019_mcan} for transformer attention on four datasets (VQAv2, VQA-CPv2, VQA-Introspect and VQA-Rephrasings). 
The standard VQAv2 and VQA-CPv2 use simple accuracy, but VQA-CPv2 has intentionally dissimilar answer distributions in the train and validation splits to not allow exploitation of priors. 
In general, low accuracy on both model architectures indicates a harder question (Equation \ref{eq:vqa_acc}).

\begin{equation}\label{eq:vqa_acc}
    \begin{split}
    &\text{score}_\text{VQA}(\textcolor{red}{\text{ans}}) = \\ &\text{score}_\text{VQA-CP}(\textcolor{red}{\text{ans}}) = \\
    &1-(\min(\frac{\text{\# of annotators that said \textcolor{red}{ans}}}{3}, 1))
    \end{split}
\end{equation}

VQA-Introspect asks additional perceptual sub-questions to VQAv2 and tests consistency w.r.t. visual grounding. 
If a model is correct on the main question "Can birds fly?", but fails the perceptual sub-question "Are the birds in the air?", it is inconsistent and the question is potentially too easy, as it can be answered from the question alone. 
We assign the (binary encoded) four combinations of "main correct/incorrect" and "all sub-questions correct/incorrect" numerical values to combine it with the other metrics. 
In Equation \ref{eq:intro}, we purposefully assign a high difficulty to a question where the perceptual sub-question is correctly answered, but the main reasoning question is not (01) and a low difficulty for a question that seems to exploit question bias (10).

\begin{equation}\label{eq:intro}
\text{score}_\text{Intro} = \left\{\begin{array}{ll} 1.0 & 00 \text{ or } 01 \\ 0.25 & 10 \\ 0.0 & 11\end{array}\right.
\end{equation}

Finally VQA-Rephrasings tests robustness against 3 linguistic variations per question and measures this with a "consensus score" -- the share of fully correctly answered subsets of size $k$ of a question and its rephrasings. 
It is unclear how to interpret different settings of $k$, so we set $k=1$, which simplifies to simple accuracy over the rephrasings (Equation \ref{eq:rep}).

\begin{equation}\label{eq:rep}
    \begin{split}
    &\text{score}_\text{Rep} = \\
    &1-(\frac{\text{\# of correct k-sized subsets}}{\text{\# of k-sized subset}}
    \end{split}
\end{equation}

We combined the four resulting scores of each MFB and MCAN in equal parts (Equation \ref{eq:diffscore}). 
Since not all our candidate stimuli are present in all four datasets used in the difficulty score, there is a set $S$ of $|S|\in[1,4]$ scores per question-image pair. 
We penalize the cases where only one score is available by normalizing over $\text{avg}(|S|)$ instead of $|S|$ to counterweight the uncertainty it brings.

\begin{equation}\label{eq:diffscore}
    \begin{split}
    &\text{difficulty} = \\
    &\frac{1}{max(|S|, \text{avg}(|S|))}\cdot\sum_{s\in S}(\frac{\text{score}_s^\text{MFB}+\text{score}_s^\text{MCAN}}{2})
    \end{split}
\end{equation}

\begin{figure}[ht]
    \centering
    \includegraphics[width=\linewidth]{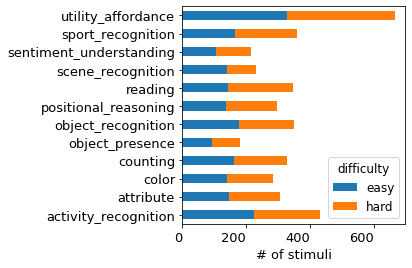}
    \caption{Final distribution of difficulty per tagged reasoning type in VQA-MHUG. Clearly some types like \textit{reading} and \textit{counting} are harder than others.}
    \label{fig:diff_in_mhug}
\end{figure}

Figure \ref{fig:diff_in_mhug} shows the resulting distribution of difficulty scores. 

\section{Significance between models}
\label{sec:test_signf_between_models}
Table \ref{tab:JSD_corr_signf} shows the significance of the differences in Rank Correlation and JSD for pairs of models.

\begin{table*}[ht]
\centering
\resizebox{\textwidth}{!}{%
\begin{tabular}{@{}lllll@{}}
\toprule
Between Model Comparison & Image Correlation   & Image JSD & Text Correlation & Text JSD            \\ \midrule
MCAN\_G vs MCAN\_R       & ***                 & ***       & ***              & ***                 \\
MCAN\_G vs PYTHIA        & ***                 & ***       & ***              & ***                 \\
MCAN\_G vs BAN           & ***                 & ***       & ***              & p\textgreater{}0.05 \\
MCAN\_G vs MFB           & ***                 & ***       & ***              & ***                 \\
MCAN\_R vs PYTHIA        & ***                 & ***       & ***              & ***                 \\
MCAN\_R vs BAN           & ***                 & ***       & ***              & p\textgreater{}0.05 \\ \midrule
PYTHIA vs BAN            & p\textgreater{}0.05 & ***       & ***              & ***                 \\
PYTHIA vs MFB            & ***                 & ***       & ***              & ***                 \\
BAN vs MFB               & ***                 & ***       & ***              & ***                
\end{tabular}%
}
\caption{We performed a paired t-test to indicate if the differences between correlation and JSD scores is statistically significant where p\textless{}0.05 (*), p\textless{}0.01 (**), p\textless{}0.001 (***). We show that for all models, the image correlation scores are statistically differ except when comparing the Pythia and BAN models. The image JSD scores and text correlation scores are significantly different for all models. The difference between models text JSD scores are significant, except for between BAN and both MCAN networks.}
\label{tab:JSD_corr_signf}
\end{table*}

\section{Experimental Setup}
\begin{figure}[ht]
    \centering
    \includegraphics[width=0.9\linewidth]{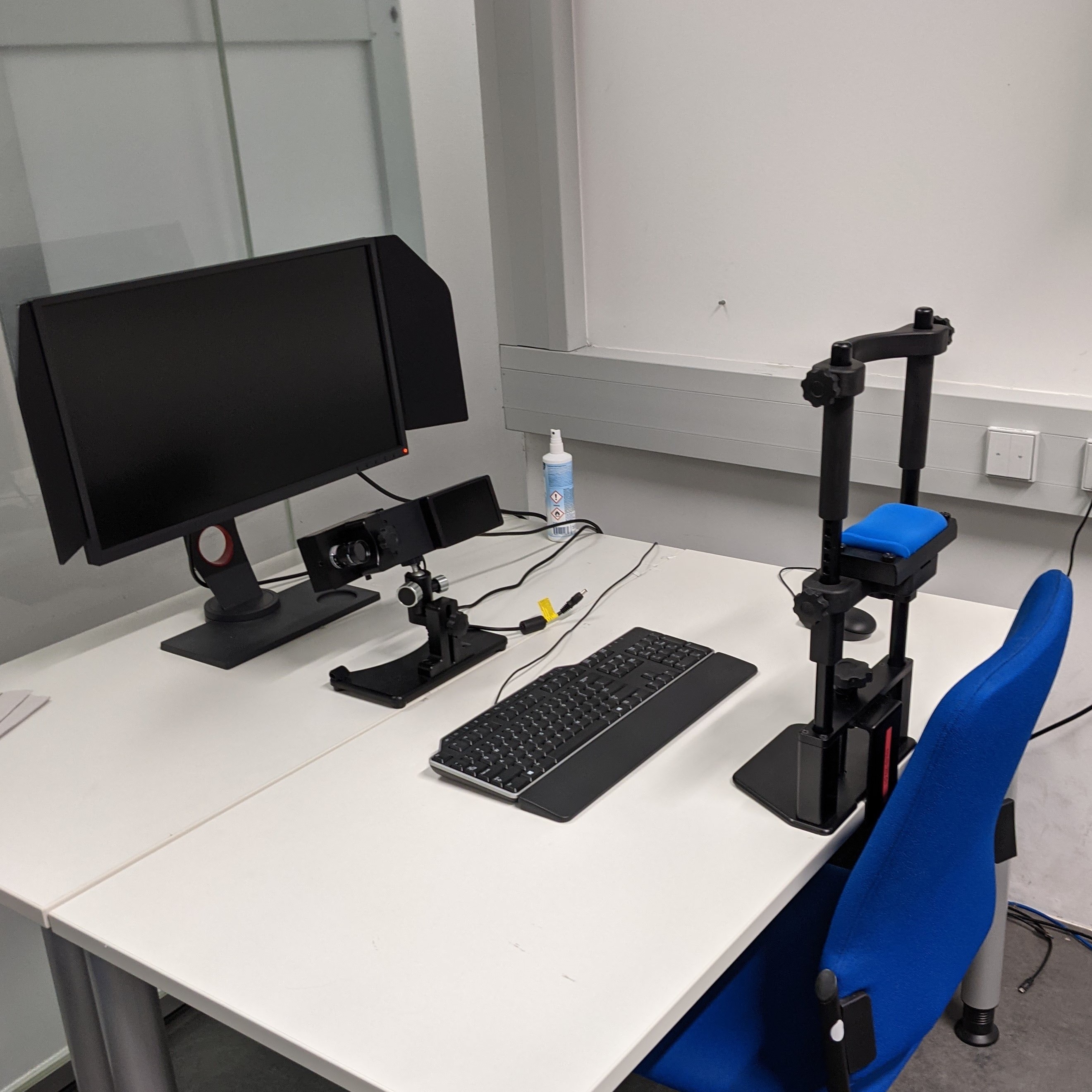}
    \caption{Setup of the eye tracker in our lab}
    \label{fig:tracker_setup}
\end{figure}

Binocular gaze data was collected with an EyeLink 1000 plus remote eye tracker at 2kHz. To ensure gaze estimation accuracy, participants were asked to use a mounted chin rest (see Figure \ref{fig:tracker_setup}). The stimuli was shown on a 24.5'' screen with resolution of $1920\times1080$ pixels. The monitor was placed 90cm in front of the participants.
\label{sec:experimental_setup}

\section{MHUG vs. AiR-D Examples}
\begin{figure}[ht]
    \centering
    \includegraphics[width=0.9\linewidth]{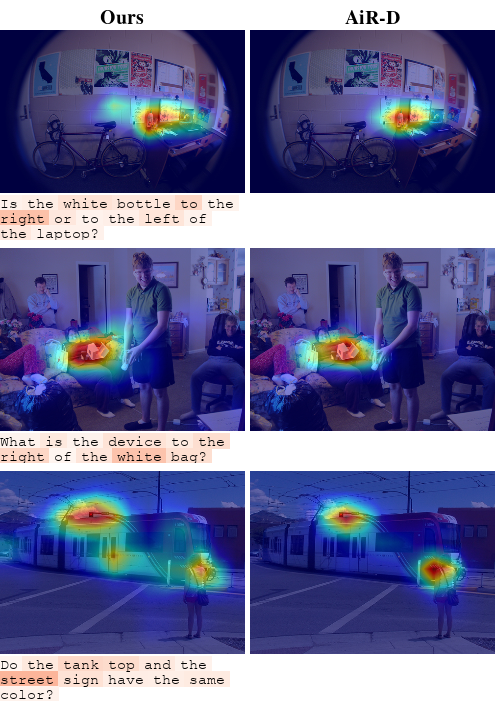}
    \caption{Examples of MHUG gaze vs. AiR-D gaze}
    \label{fig:mhug-vs-air1}
\end{figure}

\begin{figure}[ht]
    \centering
    \includegraphics[width=0.9\linewidth]{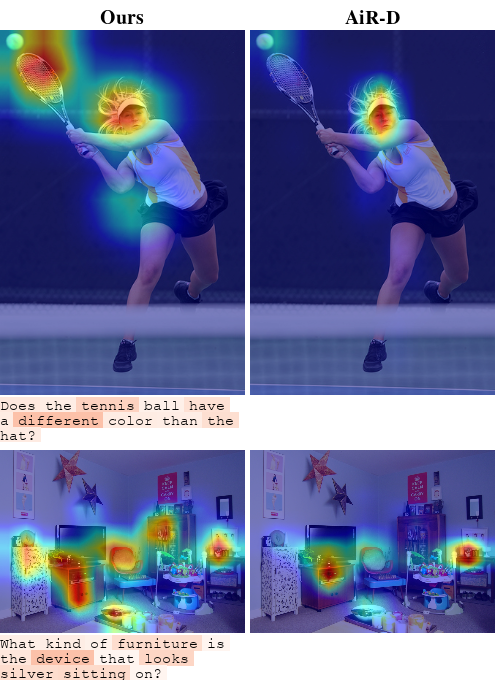}
    \caption{Examples of MHUG gaze vs. AiR-D gaze}
    \label{fig:mhug-vs-air2}
\end{figure}

The AiR-D dataset does not overlap with VQAv2, as such we separately visualize a selection of examples (see Figures \ref{fig:mhug-vs-air1} and \ref{fig:mhug-vs-air2}) from the overlapping 195 additional stimuli presented to humans during the VQA-MHUG data collection.
\label{sec:mhug-vs-air}

\section{More Model Examples}
\label{sec:more_examples}
\begin{figure*}[ht]
    \centering
    \includegraphics[width=0.9\textwidth]{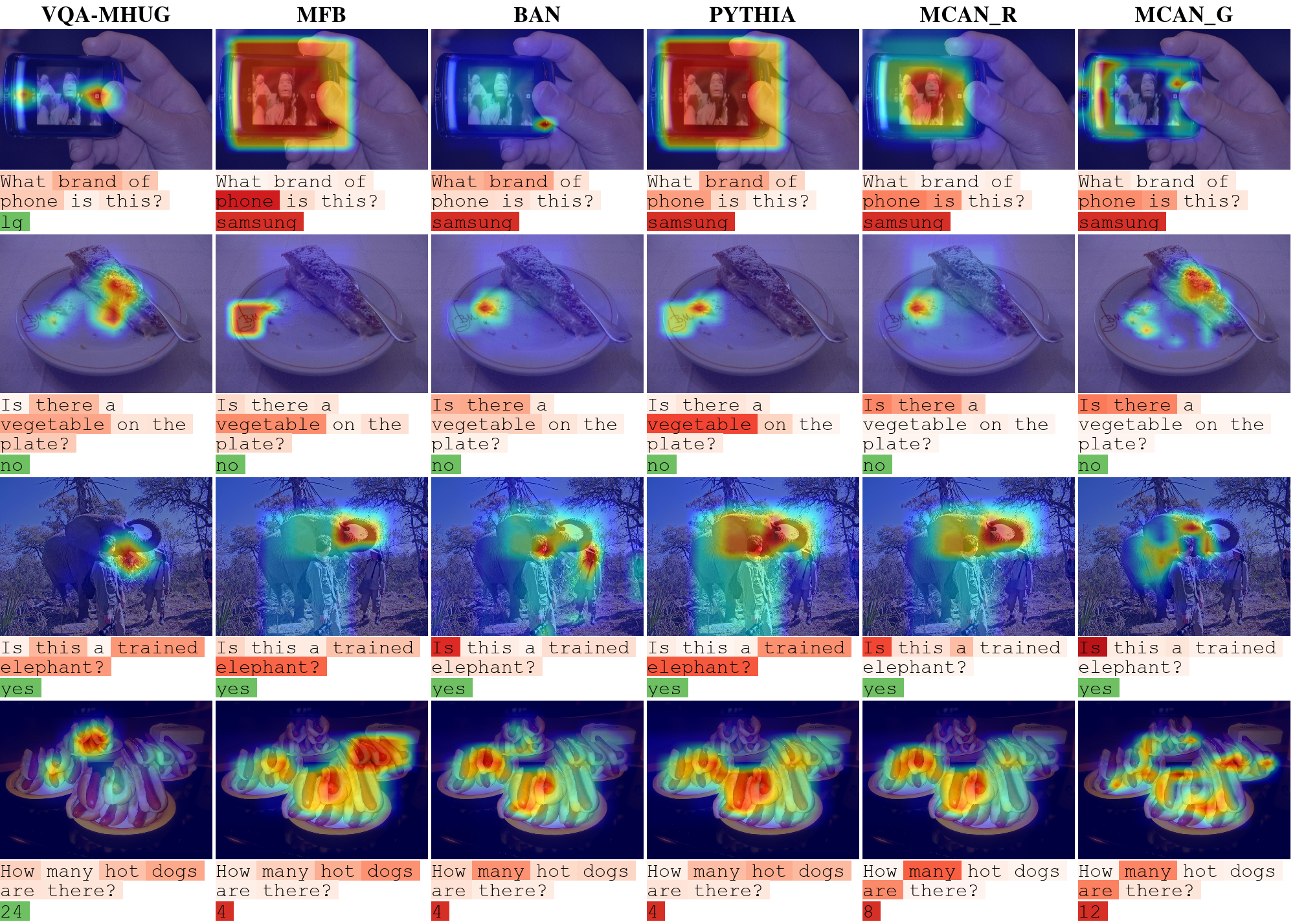}
    \caption{Comparison of VQA-MHUG attention maps and the model extracted attention maps on text and images, \textbf{where image attention correlation is high}.}
    \label{fig:high_img_corr}
\end{figure*}

\begin{figure*}[ht]
    \centering
    \includegraphics[width=0.9\textwidth]{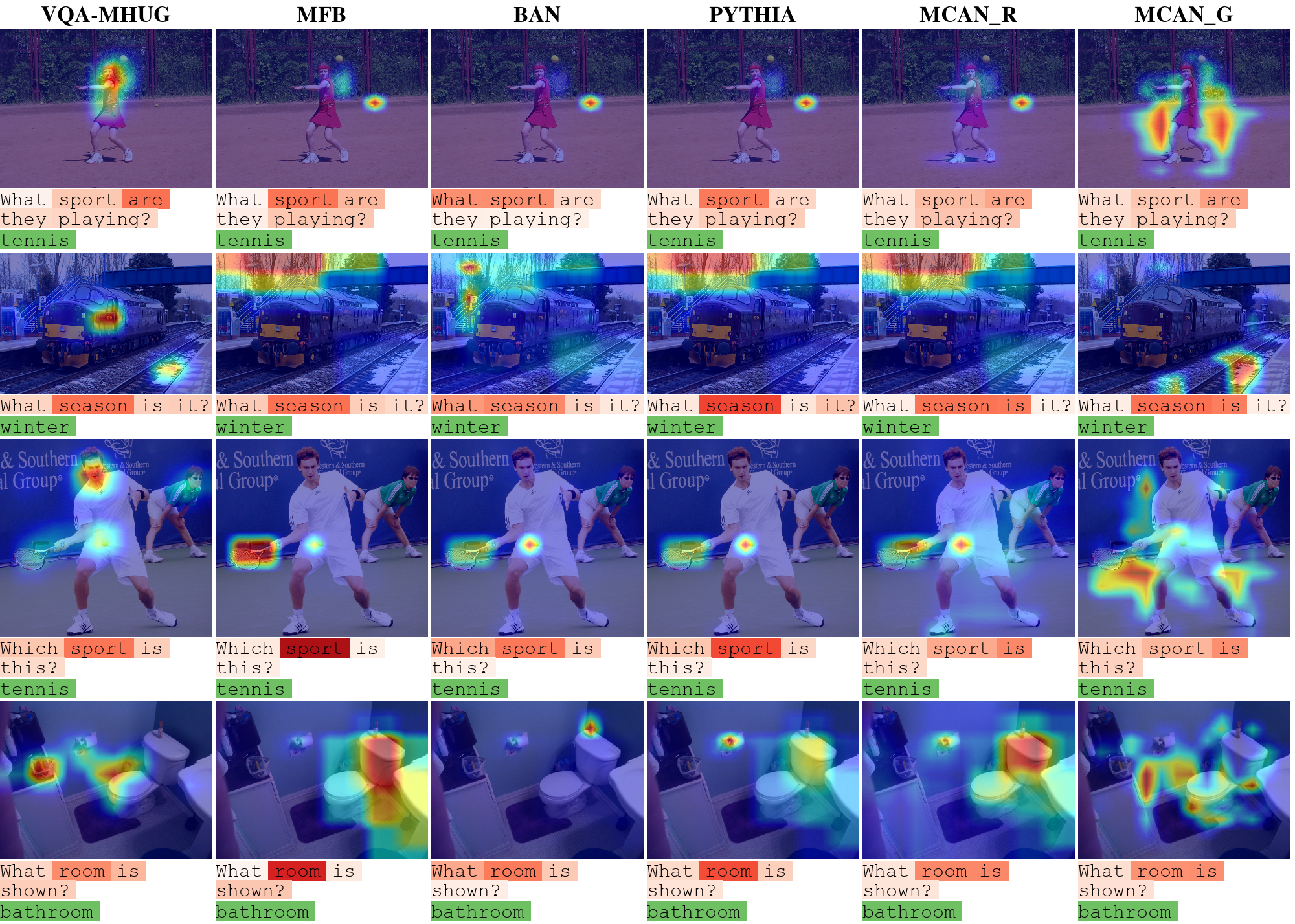}
    \caption{Comparison of VQA-MHUG attention maps and the model extracted attention maps on text and images, \textbf{where text attention correlation is high}.}
    \label{fig:high_txt_corr2}
\end{figure*}

Figures \ref{fig:high_img_corr} and \ref{fig:high_txt_corr2} show additional visualization examples of VQA-MHUG data in comparison with the extracted model data. We randomly sampled question-image pairs with high image attention correlation (Figure \ref{fig:high_img_corr}) and high text attention correlation (Figure \ref{fig:high_txt_corr2}).
